\documentclass[letterpaper, 10 pt, conference]{ieeeconf}
\IEEEoverridecommandlockouts
\usepackage{microtype}
\usepackage{graphicx}
\usepackage{subfigure}
\usepackage{booktabs} 

\usepackage{hyperref}

\usepackage{amsmath}
\usepackage{amssymb}
\usepackage{bm}
\usepackage{xcolor}

\usepackage{tikz}
\usetikzlibrary{bayesnet}

\title{\LARGE \bf
Plan-Space State Embeddings for Improved Reinforcement Learning
}

\author{Max Pflueger$^{1}$ and Gaurav S. Sukhatme$^{1}$%
\thanks{$^{1}$Department of Computer Science, University of Southern California, Los Angeles, California, USA
        {\tt\small \{pflueger, gaurav\}@usc.edu}}%
}

\begin{document}

\maketitle

\begin{abstract}

Robot control problems are often structured with a policy function that maps state values into control values, but in many dynamic problems the observed state can have a difficult to characterize relationship with useful policy actions.
In this paper we present a new method for learning state embeddings from plans or other forms of demonstrations such that the embedding space has a specified geometric relationship with the demonstrations.
We present a novel variational framework for learning these embeddings that attempts to optimize trajectory linearity in the learned embedding space.  
We show how these embedding spaces can then be used as an augmentation to the robot state in reinforcement learning problems.
We use kinodynamic planning to generate training trajectories for some example environments, and then train embedding spaces for these environments.
We show empirically that observing a system in the learned embedding space improves the performance of policy gradient reinforcement learning algorithms, particularly by reducing the variance between training runs.
Our technique is limited to environments where demonstration data is available, but places no limits on how that data is collected. 
Our embedding technique provides a way to transfer domain knowledge from existing technologies such as planning and control algorithms, into more flexible policy learning algorithms, by creating an abstract representation of the robot state with meaningful geometry.

\end{abstract}

\section{Introduction}

Recently, there has been a lot of interest in ways to solve difficult robotics problems using learning. 
Learning based approaches offer a lot of advantages in flexibility and adaptability, and don't require the engineer to offer much knowledge about the behavior or structure of the system.  Reinforcement learning (RL), a common approach here, is
used to solve a problem from the ground up, only guided by a reward function and experience interacting with the world, with no knowledge of the way the world works.  Recent advances in 
reinforcement learning
have led to significant success in solving challenging problems, but usually only in spaces where it is possible to leverage enormous data collection capabilities, either with large numbers of physical robots or simulated environments.  A lack of world knowledge makes longer horizons increasingly difficult for RL algorithms, so we are interested in looking for ways to incorporate knowledge from other sources to improve the performance of our RL learners.

Substantial world knowledge is available in the form of existing planning techniques for robots that are able to reason directly about cause and effect relationships in manipulating the world.  
In many cases they may not be able to provide fully optimized solutions to planning problems, but they can offer a significant prior base of knowledge about what useful motions in an environment look like.  We allow them to express this knowledge by providing a set of demonstration trajectories that show useful, goal directed actions in our environment.
In this work we aim to make this knowledge available to an RL or other planning agent in the form of an embedding of the state space of the robot and environment such that the geometry of the embedding space has a desirable relationship to the demonstration trajectories.

The idea of our embedding objective is that if we take a demonstrated trajectory, the path of that trajectory in the embedding space should be linear and constant velocity.  Our system pursues this objective on short overlapping segments of the demonstrations, thus driving the full embedded demonstrations towards linearity as well.
In this objective, an ideal representation would be one that creates a metric space for our system, with distances representing actual manipulation distance in our system.  It is also desirable to do this embedding in a variational framework as this will allow us to avoid the difficult problem of defining a distance metric that makes sense for our raw state space.

Although these learned embedding spaces are likely to be imperfect against our objective, we hypothesize that they provide a useful intermediate representation for RL algorithms that attempt to learn a policy function.  The mechanism here could be a reduction in the learned function distance from our embedding to an ideal target function or smoothing of the reward surface enabling more reliable convergence during training.  We test this empirically by training embedding networks for sample environments and then using standard RL algorithms to learn behaviors in these environments.  We compare the training performance of using our state embeddings against learning from the unmodified state observations and see significant improvements in training performance and reliability.

An observation we make in our experiments is that in some problems, by augmenting the observed state with our embedding space we are able to significantly reduce the variance in training performance that results from changes in the initial random seed, which is a significant confounding factor in practical RL as well as in doing good RL research.

In Section \ref{sec:psse} we cover the math behind our formulation, and then address some implementation details in Section \ref{sec:implementation}.  We will then cover the results of our experiments in Section \ref{sec:experiments}.

\section{Related Work}

Our optimization objective is inspired by work in word embeddings (such as \cite{mikolov2013efficient}) for language models where words are mapped to vectors in a latent space.  Canonical techniques in this space use sequence data from natural language and set an optimization objective that words that occur sequentially in language should have embeddings that are nearly co-linear in the latent space.  We treat the sequence data from a robot trajectory in a similar way, and develop a new embedding formulation to meet the needs of our robot state embeddings, which we discuss further in Section \ref{sec:psse}.

Robot learning from demonstrations exists in multiple other forms.  Inverse reinforcement learning supposes that the demonstrations are drawn from a distribution optimizing some reward function, and attempts to learn that reward function \cite{ng2000algorithms}\cite{ziebart2008maximum}\cite{pflueger2019rover}.  A very similar problem form known as imitation learning or behavior cloning aims to learn to reproduce the behavior of the demonstrator, without explicitly learning the underlying reward function.  Generative adversarial imitation learning (GAIL) \cite{ho2016generative}\cite{finn2016connection} combines these ideas with inspiration from generative adversarial networks \cite{goodfellow2014generative} by training a discriminative cost function simultaneously with an RL agent.
Our approach exists slightly outside these techniques in that while we assume the demonstrator has some form of optimality in movement costs, we do not attempt to learn the demonstrator's reward function per se, and indeed in our RL step we may drop in a reward function of our choice.  

Embedding spaces have been applied before in reinforcement learning, such as in the form of action space embeddings to support skill transfer between tasks \cite{hausman2018learning}. 
State embeddings, while less common in RL, have been used in robotics applications.  
In \cite{ichter2019robot}, Ichter and Pavone construct a way to train embedded state spaces with predictable dynamics in the latent space to enable sampling based planning techniques to function directly in the latent space.
An important point of comparison for our work is the embed-to-control algorithm proposed in \cite{watter2015embed}.  This algorithm uses a variational model to learn state embeddings supporting the objective of having locally linear state transition dynamics in the embedding space.  
Our approach also takes the high-level form of a variational auto-encoder \cite{kingma2013auto}, with some specific structure in the distribution of states and embedded states.
Although the embed-to-control structure and objective is different from our own, they use some very similar mathematical tools for resolving underlying difficulties and learning the embedding.

\section{Plan-Space State Embedding}
\label{sec:psse}

In this section we will describe our formulation for learning a plan-space state embedding.  This embedding space is built on the concept of constraining demonstrated trajectories to be linear in the embedding space.  We present an initial direct formulation of this objective, then show why optimizing that objective may be problematic, and fix those issues with a more sophisticated variational model.

\subsection{Trajectory Linearity Model}

We wish to learn an encoding of the raw state space such that locality in the embedding implies locality in manipulation or control distance.  To do this we take advantage of the availability of a dataset of demonstration trajectories believed to be efficient, or ideally near optimal, and { due to the optimal substructure property, we} note this would also imply that snippets of these trajectories are efficient for their respective sub-goals {(final position in the snippet)}. 
To satisfy our embedding objective, we attempt to constrain the demonstrated trajectories to form straight lines through our embedding space.  We form a loss function on this by sampling evenly spaced triplets of states along a demonstrated trajectory, and rewarding the middle point for having an embedding value close to the average of the embeddings of the edge points.

{
\subsection{Preliminaries}

We define our robot state as $x\in X$, and we define our embedding space as $Z= \mathbb{R}^n$.  The dimensionality $n$ of the embedding space will be a hyperparameter, and can in general be greater or less than the robot state dimensionality.  
We consider a trajectory to be a vector of states such as $x_0,...,x_f$, and sampled triplets from trajectories are notated as $\bm{x} = (x_{t-1}, x_t, x_{t+1})$, although in general the states may be more than a single time step apart from each other.
We also define the trajectory distribution of our training data as $\tau(\bm{x})$.
}

\subsection{Direct State Encoding}

A simple and direct form of this problem would be to use a modified auto-encoder structure. 
We define two parametric functions $h^{\textup{enc}}_{\phi}:X \mapsto Z$ and $h^{\textup{dec}}_{\theta}:Z \mapsto X$, for encoding and decoding respectively.  We could then define our loss function as 

{
\begin{equation}
    \mathbb{E}_{
    \left.\begin{matrix}x_{t-1}\\ x_t\\ x_{t+1}\end{matrix} \right\}  \sim \tau} 
    \left\Vert
    x_t - 
    h^{\textup{dec}}_{\theta} \left(\frac{1}{2}(h^{\textup{enc}}_{\phi}(x_{t-1}) + h^{\textup{enc}}_{\phi}(x_{t+1})\right) 
    \right\Vert_2
\end{equation}
This defines our loss in terms of the $\ell^2$-norm of the difference in state values, however it is not clear that this is a reasonable distance metric in the robot state space, particularly for spaces that include velocities or higher order inertial states. We could make this more general by allowing an arbitrary distance metric $d(x,y)$ in the state space as
}

\begin{equation}
    \mathbb{E}_{
    \left.\begin{matrix}x_{t-1}\\ x_t\\ x_{t+1}\end{matrix} \right\}  \sim \tau} 
    \left[ 
    d\left(
    x_t, 
    h^{\textup{dec}}_{\theta} \left(\frac{1}{2}(h^{\textup{enc}}_{\phi}(x_{t-1}) + h^{\textup{enc}}_{\phi}(x_{t+1})\right) 
    \right) 
    \right]
\end{equation}
Unfortunately, in this formulation $d$ becomes a powerful hyperparameter without any obviously good choices.  
In another sense, if we had access to a good distance metric on our space, we would already have a solution to the majority of our problem.  In the next section, we propose a probabilistic approach that can be optimized without the need for a distance metric in the robot state space.

\subsection{Variational State Encoding}

\begin{figure}
    \centering
    \tikz{ %
        \node[latent] (z1) {$z_{t-1}$}; %
        \node[latent, right=of z1] (z2) {$z_{t}$}; %
        \node[latent, right=of z2] (z3) {$z_{t+1}$}; %
        \node[obs, below=of z1] (x1) {$x_{t-1}$}; %
        \node[obs, below=of z2] (x2) {$x_{t}$}; %
        \node[obs, below=of z3] (x3) {$x_{t+1}$}; %
        \plate[] {plate1} {(z1) (z2) (z3) (x1) (x2) (x3)} {N}; %
        \edge{z1} {z2}; %
        \edge{z3} {z2}; %
        \edge{z1} {x1}; %
        \edge{z2} {x2}; %
        \edge{z3} {x3}; %
    }
    \caption{Our model of path data.}
    \label{fig:model}
    \vspace{-5pt}
\end{figure}

\begin{figure*}
\begin{center}
\includegraphics{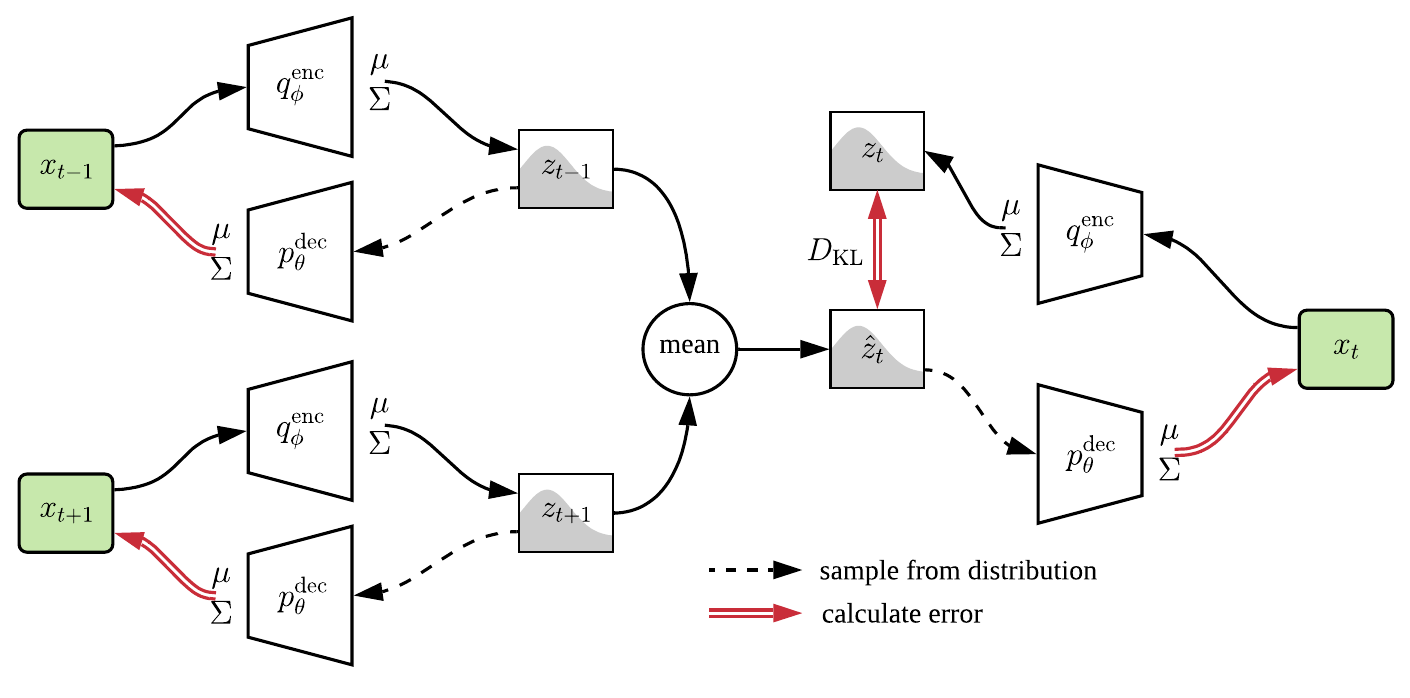}
\caption{Information flow diagram for our variational state embedding algorithm.  The green boxes with $x$ values represent samples from our training dataset.  The $z$ values are stored internally by their mean and variance, and sampled when necessary along the dotted lines.  The double red arrows represent points where terms from our evidence lower bound are calculated.  Not shown is the calculation of KL divergence of $z$ values from the embedding space prior.}
\label{psse_info_flow}
\end{center}
\end{figure*}

Instead of modeling our embedding function as a deterministic function of the state $x$, we make the embedding probabilistic, 
so $h^{enc}_{\phi}(x)=z \sim q_\phi(z|x)$. 
In this way we will now be able to more flexibly model uncertainties in paths, as well as no longer having to specify a distance metric in the state space $X$.

We assume that our encoded states $z$ are drawn from a prior distribution $z \sim p(z) = \mathcal{N}(0,1)$ and robot states $x$ are conditioned on the latent states as $x \sim p(x|z)$.  When considering a trajectory triplet $(x_{t-1}, x_t, x_{t+1})$ we say that states 
$x_t \sim p(x_t|z_t)$ for $z_{t-1}, z_t, z_{t+1}$ respectively.  However, while $z_{t-1}, z_{t+1} \sim p(z)$, $z_t$ is defined differently as $z_t = \frac{z_{t-1}+z_{t+1}}{2}$.
We create an approximation of the encoding distribution $q_\phi(z|x)$.
We also create an approximate inverse decoding distribution $p_\theta(x|z)$.

We see that when we define $\bm{x} = (x_{t-1}, x_t, x_{t+1})$ and $\bm{z} = (z_{t-1}, z_{t+1})$ our problem takes the form of a variational auto-encoder \cite{kingma2013auto}, where we wish to maximize the marginal log likelihood of the data:

\begin{equation}
    \log p_\theta(\bm{x}) = D_{KL}(q_\phi(\bm{z}|\bm{x})||p_\theta(\bm{z}|\bm{x})) + \mathcal{L}(\theta,\phi;\bm{x})
\end{equation}

As with other variational methods (\cite{kingma2013auto, watter2015embed}), instead of optimizing this objective directly we optimize the variational lower bound $\mathcal{L}$, expanded below:

\begin{align}
    \mathcal{L}(\theta,\phi;\bm{x}) & = \mathbb{E}_{\bm{z}\sim q_\phi(\bm{z}|\bm{x})}\left[ -\log q_\phi(\bm{z}|\bm{x}) + \log p_\theta(\bm{x},\bm{z}) \right] \\
    \begin{split} 
        & = -D_{KL}\left( q_\phi(\bm{z}|\bm{x}) || p_\theta(\bm{z}) \right) \\
        &\quad + \mathbb{E}_{\bm{z}\sim q_\phi(\bm{z}|\bm{x})}\left[ \log p_\theta(\bm{x}|\bm{z}) \right]
    \end{split}
\end{align}

We choose to restrict $p$ and $q$ to the space of multivariate normal distributions with diagonal covariance matrices.  They are parameterized with neural networks which produce $\mu$ and $\Sigma$ as a function of $z$ and $x$ for $p$ and $q$ respectively. In this form we can use the generic stochastic gradient variational bayes (SGVB) estimator from \cite{kingma2013auto}.  If we further approximate $q_\phi(\bm{z}|\bm{x})$ as

\begin{equation}
    \tilde{q}_\phi(\bm{z}|\bm{x}) = q_\phi(z_{t-1}|x_{t-1})q_\phi(z_{t+1}|x_{t+1})
\end{equation}
then we can analytically calculate the KL-divergence and use the lower variance form of the SGVB estimator.  Unfortunately, in this case we have unbound a key constraint in our system, so we add an additional KL-divergence term between $\tilde{q}_\phi(z_t|\bm{x})$ and $q_\phi(z_t|x_t)$, modifying the lower bound as follows:

\begin{equation}
\label{math:objective}
\begin{split}
    \mathcal{\hat{L}} = 
      & -D_{KL}\left( \tilde{q}_\phi(\bm{z}|\bm{x}) || p_\theta(\bm{z}) \right) \\ 
      & + \mathbb{E}_{\bm{z}\sim \tilde{q}_\phi(\bm{z}|\bm{x})}\left[ \log p_\theta(\bm{x}|\bm{z}) \right] \\
      & - \lambda D_{KL}(\tilde{q}_\phi(z_t|\bm{x}) || q_\phi(z_t|x_t))
\end{split}
\end{equation}

where $\lambda > 0$ is a training hyperparameter.  These calculations are visualized in the information flow diagram Figure~\ref{psse_info_flow}.  Note that in this diagram, $z$ values are shown where we internally track the distribution of the respective $z$'s, and these distributions are only sampled when necessary (shown as dotted lines).

We note that $\tilde{q}_\phi(z_t|\bm{x})$ is a distribution over $z_t$ and we will refer to values sampled from this distribution as $\hat{z_t}$.  Because we have assumed that our $z$'s are drawn from a normal distribution, we can analytically calculate the distribution $\tilde{q}_\phi(z_t|\bm{x})$ where $z_t$ is constrained to be the mean value of $z_{t-1}$ and $z_{t+1}$.  Our embedding distribution $q_\phi$ gives us 
\begin{equation}
\begin{split}
    z_{t-1} & \sim \mathcal{N}(\mu_a, \sigma_a^2) \\
    z_{t+1} & \sim \mathcal{N}(\mu_b, \sigma_b^2)
\end{split}
\end{equation}
where $\mu$ and $\sigma$ are vectors in the dimensionality of our embedding space.  The distribution of $\hat{z_t} = \frac{z_{t-1}+z_{t+1}}{2}$ is also a normal random variable as follows:

\begin{equation}
\label{math:zthat}
    \hat{z_t} \sim \mathcal{N} \left( \frac{\mu_a+\mu_b}{2}, \frac{\sigma_a^2 + \sigma_b^2}{4} \right)
\end{equation}

\subsection{Evaluation} \label{sec:eval_metric}

While we optimize the modified lower bound above, we will track performance of the embedding against a metric designed to measure whether the embedding space is creating the geometry we desire.
We look at $\ell^2$ distances in the embedding space compared with the trajectory distance between points along a trajectory.  To do this analysis we have to account for the fact that there could be arbitrary scale differences between distances in the two spaces, so we have to find a scale factor to match the embedding space distances to the trajectory space distances.  
We calculate the distance integral along each of our demonstration trajectories denoted $y_i$ and with a current snapshot of the embedding function $q_\phi$ calculate embedded distances $d_i = ||z_i^{init} - z_i^{final}||$ where $z_i$ is the mean value from $q_\phi(z_i|x_i)$.  For consistency we normalize the trajectory lengths $y_i$ by dividing them by the mean trajectory length.

We then calculate the best matching scale factor $C$ by minimizing the following error term:

\begin{equation}
    \sum_{i}(y_i-Cd_i)^2
\end{equation}

which can be calculated analytically as:

\begin{equation}
    C=\frac{\sum{d_i y_i}}{\sum{d_i^2}}
\end{equation}

We can then look at the error between the scaled embedded distances $Cd_i$ and the demonstrated path distances $y_i$.  During training of the state embedding we track the mean and standard deviation of the absolute values of these errors on our training dataset.

\section{Implementation}
\label{sec:implementation}

\begin{figure}
\begin{center}
\includegraphics{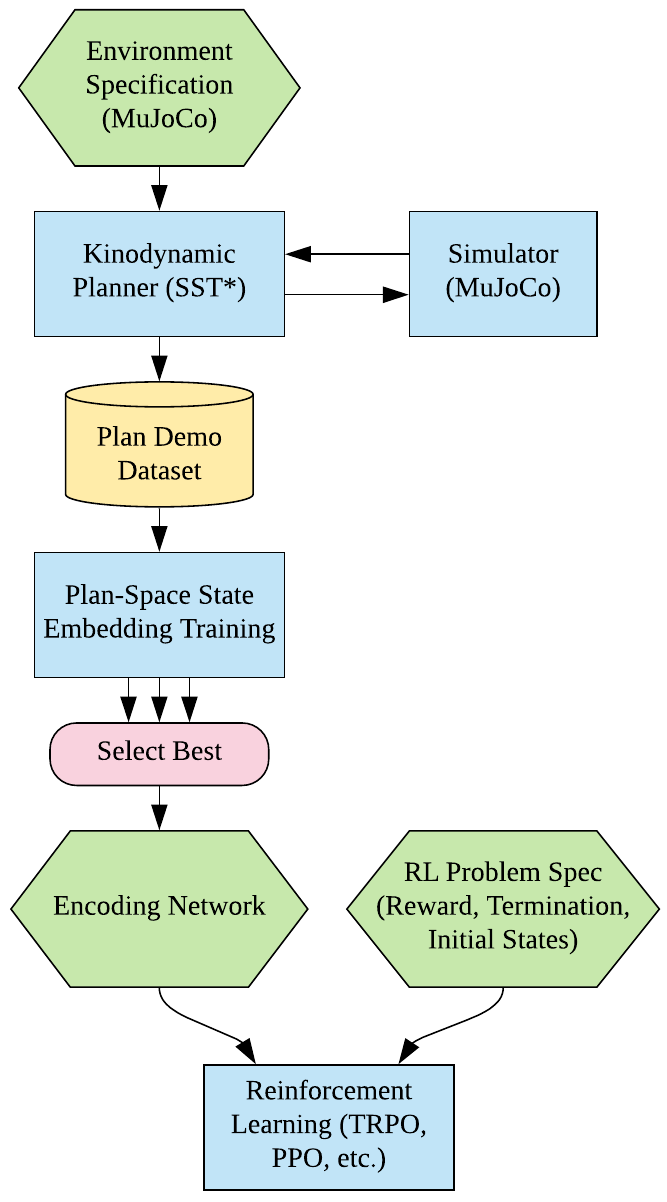}
\caption{This system diagram shows the workflow described in this paper.  We use a kinodynamic planner linked with MuJoCo environment simulation to generate a dataset of demonstration plans in our environment.  The plan-space state embedding is trained based on that dataset multiple times, and the best resulting encoding is selected to augment the observed state in reinforcement learning problems in our environment.}
\label{fig:psse_system_diagram}
\end{center}
\end{figure}

\subsection{Kinodynamic Planner Demonstrations}
Our embedding algorithm relies on a dataset of trajectory demonstrations, however it is agnostic of the source of these trajectories.  As such, we might use any available planning algorithm, expert system, or even human demonstrations to gather this data.  In this work we use the SST* algorithm \cite{li2016asymptotically} implemented in the OMPL motion planning library \cite{sucan2012the-open-motion-planning-library}.  Our robot environments are specified in and simulated by the MuJoCo robot simulator \cite{todorov2012mujoco}.

To produce the necessary trajectory demonstrations for our algorithm, we have built an interface layer between MuJoCo and OMPL.  Our interface allows us to use MuJoCo in the inner loop of OMPL algorithms as the state propagator, enabling OMPL's kinodynamic planning algorithms to function on a wide variety of spaces specified for MuJoCo with minimal additional problem specification.  We are sharing this interface code for academic use at the link below\footnote{\url{https://github.com/mpflueger/mujoco-ompl}}.

\subsection{Variational State Embedding}

We implemented the variational form of our state encoder and are sharing our source code\footnote{\url{https://github.com/mpflueger/plan-space-state-embedding}}.
Our implementation maximizes the lower variance form of the SGVB estimator using the Adam algorithm \cite{Kingma2014AdamAM}.  

For the experiments in this paper we parameterized the encoder and decoder networks as fully connected nets with 2 hidden layers of 64 units per layers.  The networks produce two sets of output for $\mu$ and the diagonal terms of the covariance matrix $\Sigma$.  The $\mu$ units have linear output.  In order to avoid numerical issues we define our $\Sigma$ to output standard deviations rather than variances, and bound outputs to the range $(0,\infty)$ by using the activation function:

\begin{equation}
    f(x) = \left\{\begin{array}{ll}
        x+1 & x\geq0\\
        exp(x) & x<0
        \end{array}\right.
\end{equation}

We also observe that the distribution of $\hat{z_t}$ is systematically smaller than the distributions of $z$ otherwise predicted by $q_\phi$ as a result of it being distributed as the average of $z_{t-1}$ and $z_{t+1}$. 
This could create a systematic error in learning because in equation \ref{math:objective} we place a cost on the distributional match of $\tilde{q}_\phi(z_t|\bm{x})$ to $q_\phi(z_t|x_t)$, despite them having the same underlying parameters with $x$'s drawn from the same distribution.  
By looking at equation \ref{math:zthat} we can see that the variance of $\hat{z_t} \sim \tilde{q}_\phi(z_t|\bm{x})$ is expected to be half that of the variance of $z \sim q_\phi(z|x)$, yet we are constraining them to come from the same distribution.  
We correct this issue by doubling the variance of $\hat{z_t}$ to match the distribution from $q_\phi(z|x)$.
Lambda ($\lambda$) from equation \ref{math:objective} was set to 0.5.

Before training begins the dataset of demo trajectories is transformed into a set of state triplets.  The triplets have even and varied spacing along the trajectories, we used step differences of $[1, 3, 5, 10, 30]$.

Figure \ref{fig:psse_training} shows the training curves for our plan-space state embedding, using 5 seeds in each configuration.  We select the best trained model for evaluation in RL environments based on performance on the metric defined in Section \ref{sec:eval_metric}.

\subsection{RL in Embedding Space}
To evaluate how using this embedding space affects the performance of reinforcement learning (RL) algorithms, we have chosen a couple of benchmark RL environments where we can modify the observed state space sent to the RL algorithm while keeping other aspects of the system the same.  To facilitate this we used implementations of trust region policy optimization (TRPO) \cite{schulman2015trust} and proximal policy optimization (PPO) \cite{schulman2017proximal} provided by the garage library \cite{garage}, and created custom environment wrappers using the pre-trained state encoders from our variational state embedding. 

Our environment wrappers always return the mean value as the embedded state.  Additionally, we can look at the effect of completely replacing the normal state space, or augmenting it by appending our embedded state vector onto the existing state vector.

\section{Experiments}
\label{sec:experiments}

\subsection{Environments}

In our experiments we used two primary environments, each with two variants.  Our environments are based on standard environments from the OpenAI Gym project, however in each case we have also created modified environments to make the tasks more challenging.

The cartpole environment has a cart that can move along a 1 meter track and has an uncontrolled spinning pole attached to it. The cartpole environment has a 4 dimensional state, including the position and velocity of both the cart and pole.  Control of the cart along the track is done by choosing an applied force (continuous value), with an upper limit on the available force.  Two RL tasks are defined for the cartpole environment.  
In the first environment the cart starts near the center of the track with the pole nearly upright and reward is given for keeping the pole upright as long as possible.  The second, much more difficult task, starts the cart and pole at random locations with small random velocities and reward is given for swinging the pole into an upright position and keeping it there.  We refer to this variant as the cartpole swingup task.

The reacher environment consists of a 2-link robot arm operating in a plane, with an objective to move the tip of the second link to a particular position.  The target position is encoded as part of the state space of the environment as the desired $x,y$ position of the end point of the robot arm.  The robot is controlled by choosing bounded joint torques.  
Our two reacher variants differ in the observation space presented to the policy function.  The initial version from Gym observes sine and cosine of the joint angles, which we refer to as reacher trig.  We also created a version of the task that observes the joint angles directly as radians instead (reacher raw).  Both versions observe joint velocities.  In both cases, the state input to our embedding function is the raw joint angles and joint velocities (a 4 dimensional space).  Note here that our state embedding does not observe the target location, so although we include some performance curves of the embedding replacing the normal observation space, the target is not observable in these cases and goal directed behavior is not possible. 

\subsection{Embedding Performance}

\begin{figure*}
\begin{center}
\includegraphics[width=0.5\linewidth]{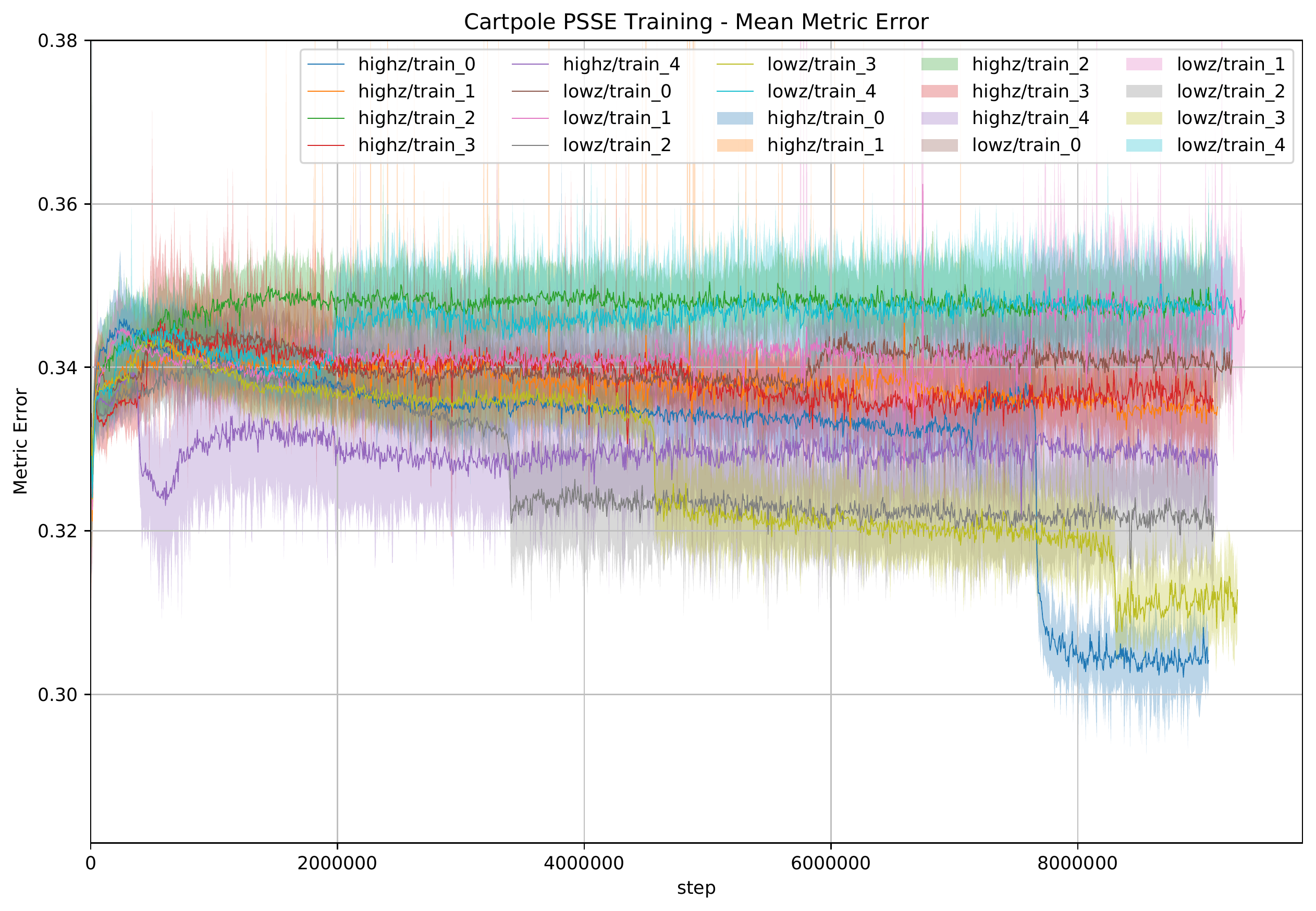}%
\includegraphics[width=0.5\linewidth]{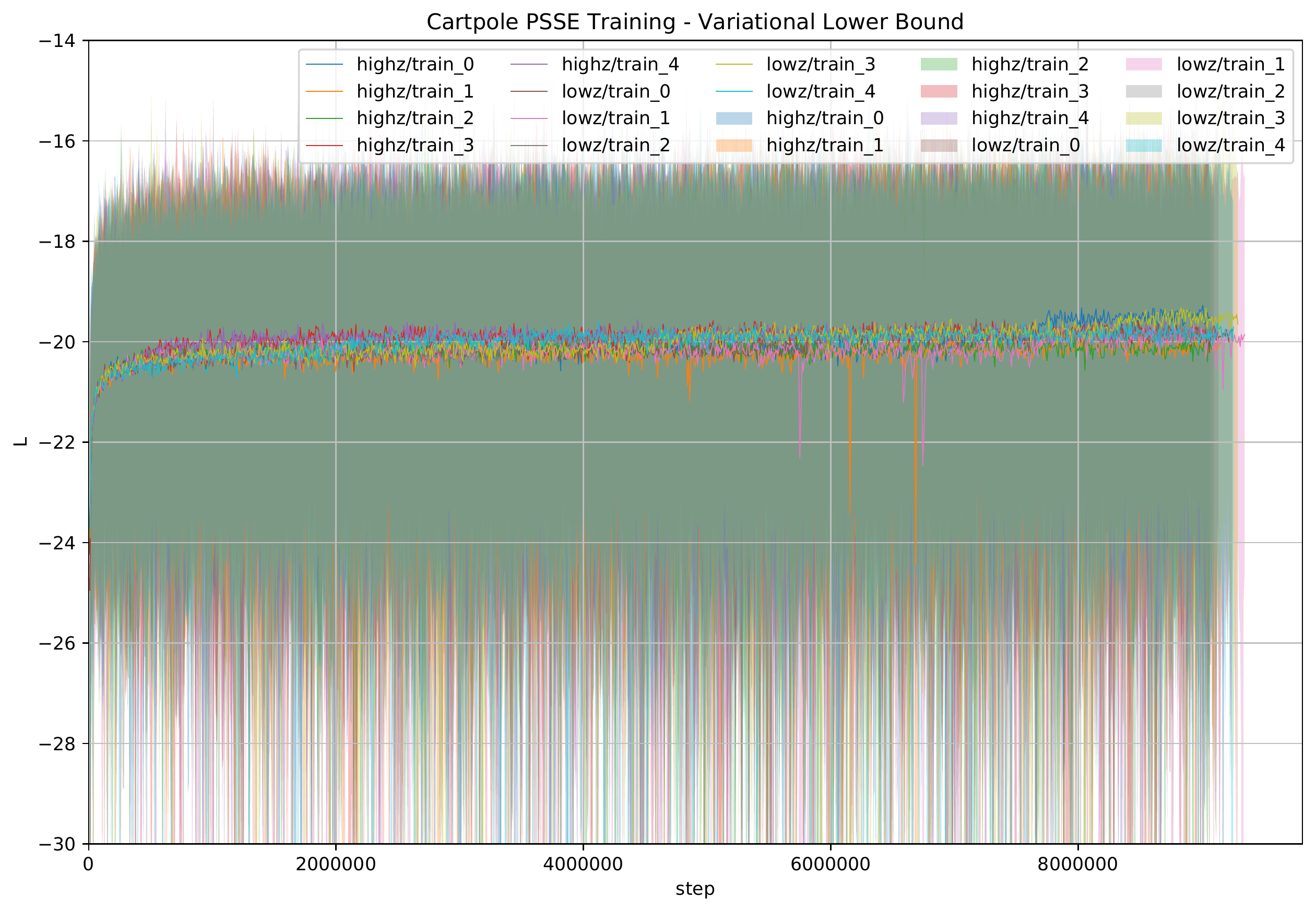}
\includegraphics[width=0.5\linewidth]{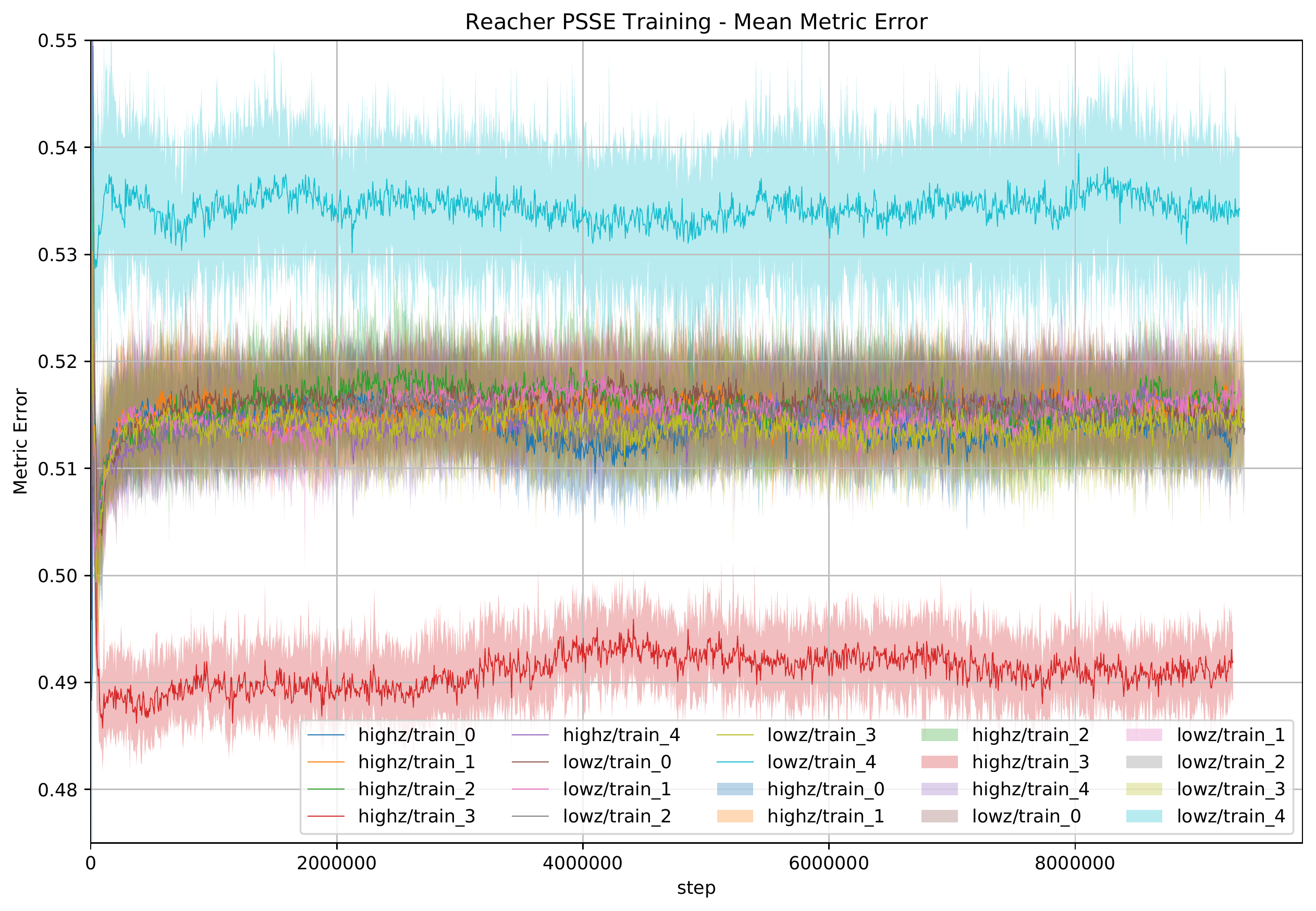}%
\includegraphics[width=0.5\linewidth]{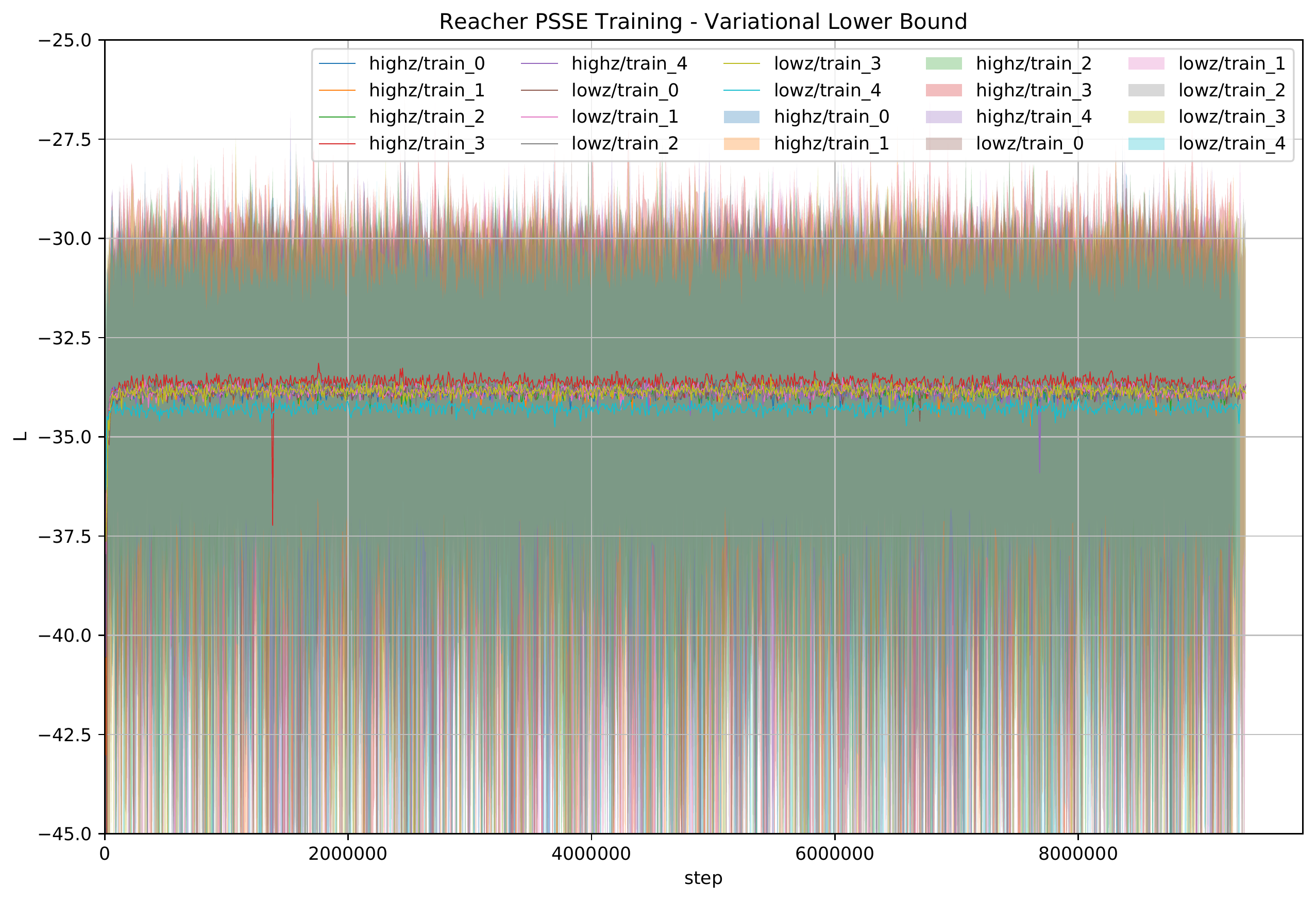}
\caption{Training curves for our plan-space state embedding in the cartpole and reacher environments.  Solid curves show a rolling window average with shaded regions showing the range of the unsmoothed curve. `highz' curves are training a 10 dimensional embedding space, `lowz' represents a 3 dimensional embedding space.  
The curves on the left show performance on the embedding performance metric described in Section \ref{sec:eval_metric}.  The curves on the right show performance on the variational lower bound described in equation \ref{math:objective}. 
Each curve shows training for a separate seed value, and the seed that achieves the best performance on the embedding metric (left) is selected for testing with RL algorithms.  The variational lower bound converges noisily but consistently, but we can see there is significant seed variation on the left.}
\label{fig:psse_training}
\end{center}
\end{figure*}

In Figure \ref{fig:psse_training} we show training curves for our plan-space state embeddings for the cartpole and reacher environments.  The plots show performance on our evaluation metric from Section \ref{sec:eval_metric} and the variational lower bound $\mathcal{\hat{L}}$ from equation \ref{math:objective}.
For each environment we train with our embedding space Z-dimension set to 10 (the high-Z configuration) and set to 3 (the low-Z configuration).  Note that both the cartpole and reacher environments have a native dimensionality of 4.  In each configuration we run training 5 times with different random seeds, and choose as our final embeddings for each configuration the one with the best performance as measured by our metric from Section \ref{sec:eval_metric}.  

We see from training curves for our state embeddings that although loss on the optimization objective $\mathcal{\hat{L}}$ converges consistently, there is still significant seed noise in the performance on our evaluation metric error.  Because this is an offline process it is easy to deal with this seed noise by training multiple seeds in parallel and selecting the best one, however, we believe that in future work it may make sense to more closely examine the convergence characteristics of this algorithm in order to get more consistent performance.

\subsection{RL Performance}

\begin{figure*}
\begin{center}
\includegraphics[width=0.5\linewidth]{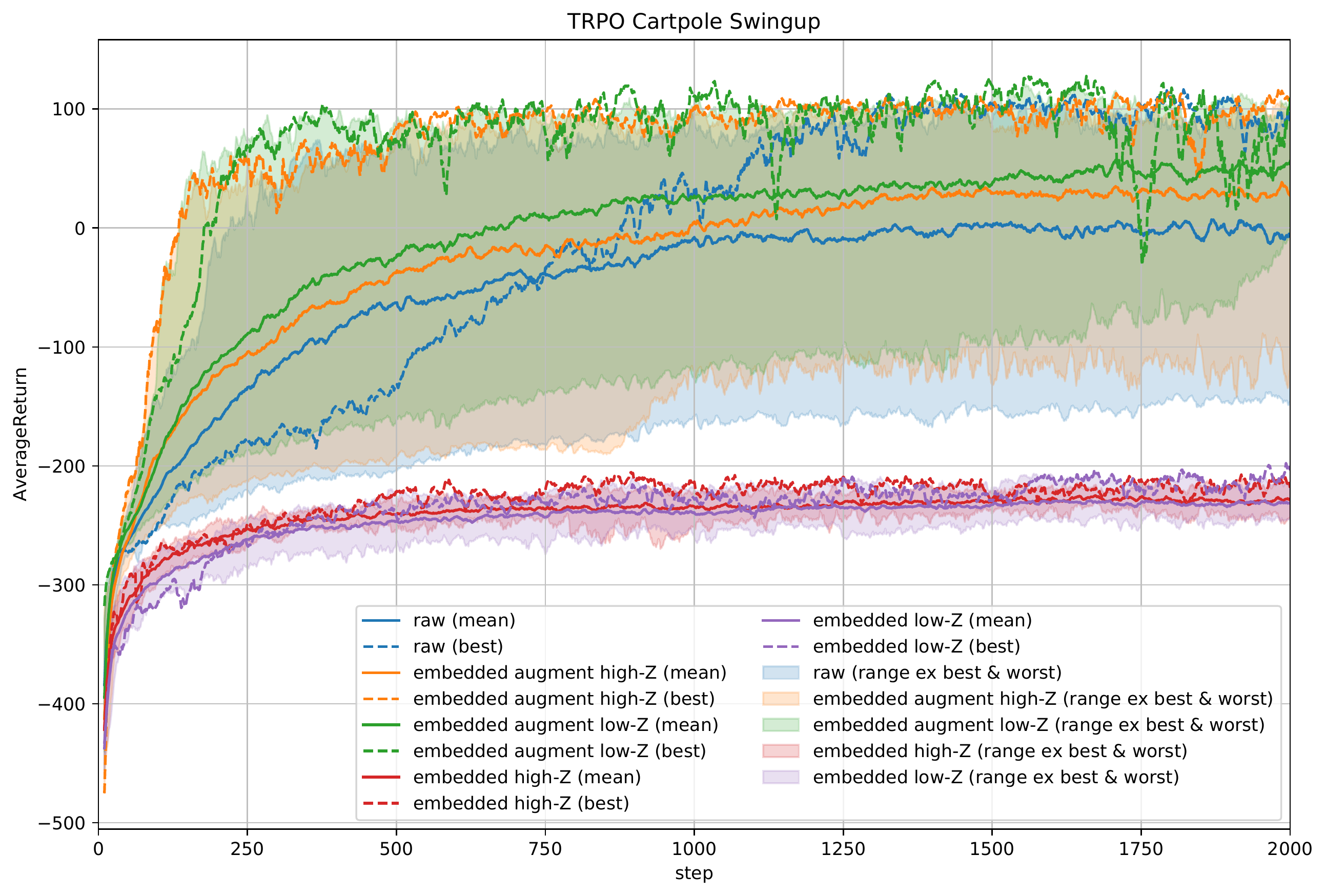}%
\includegraphics[width=0.5\linewidth]{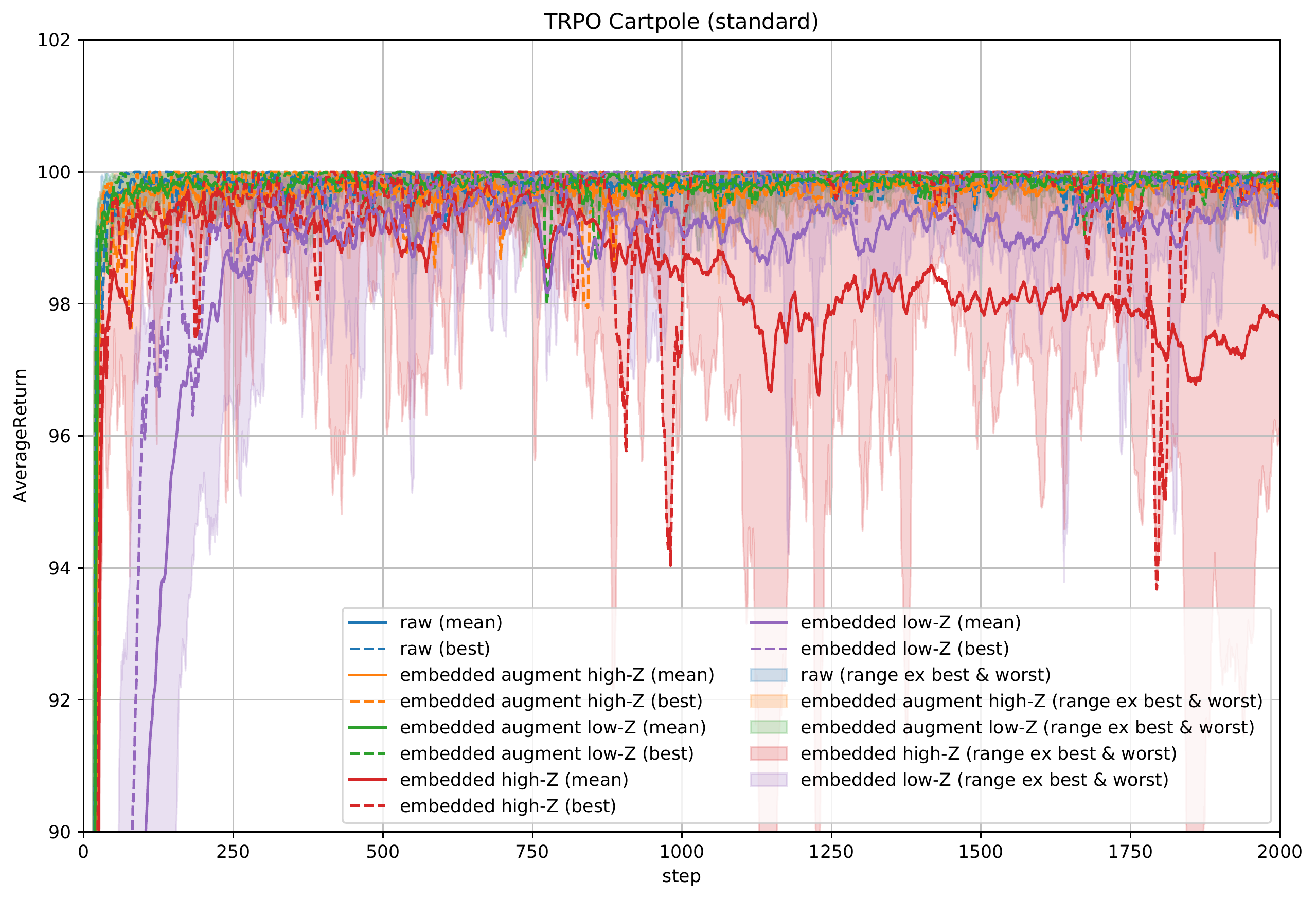}
\includegraphics[width=0.5\linewidth]{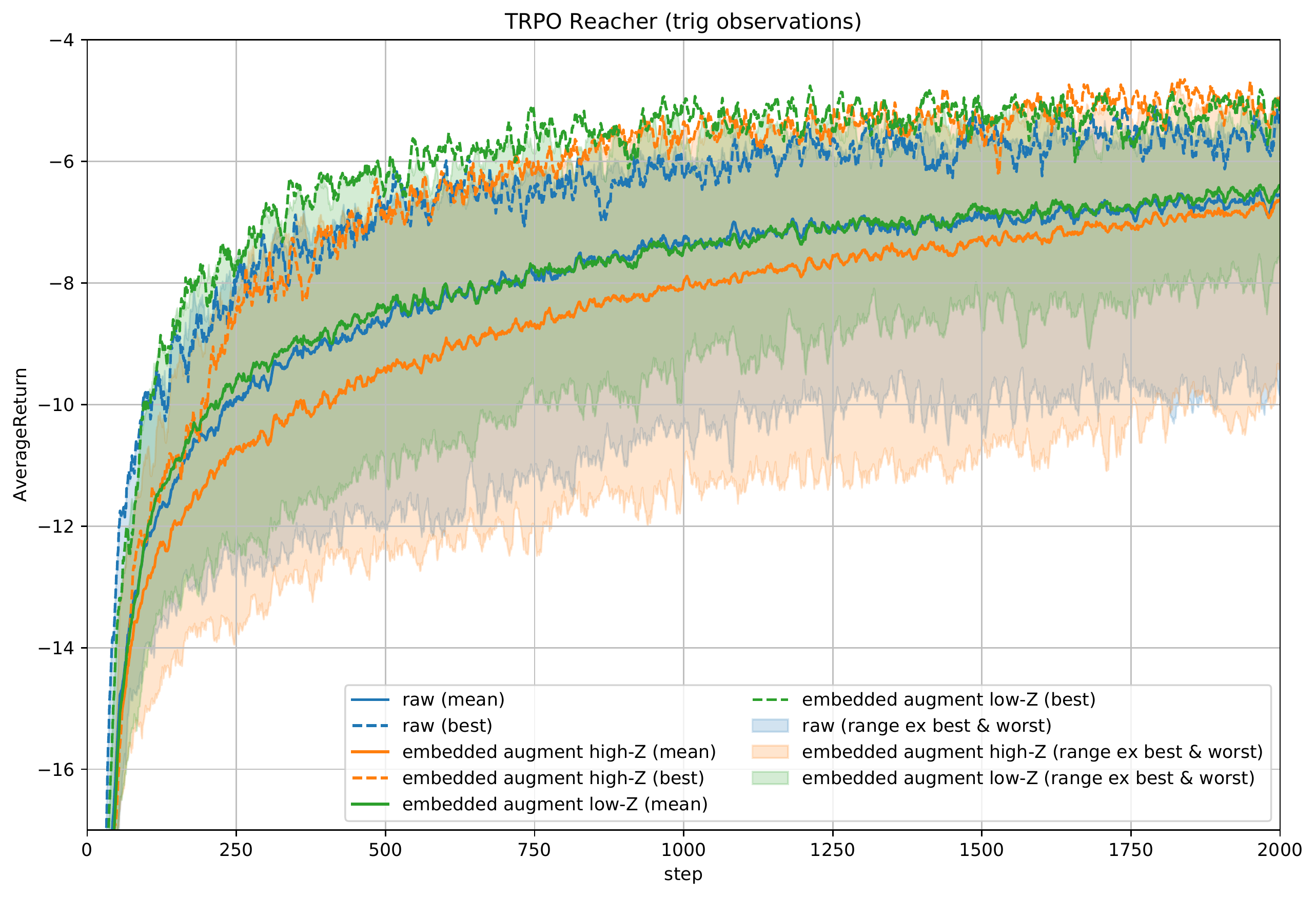}%
\includegraphics[width=0.5\linewidth]{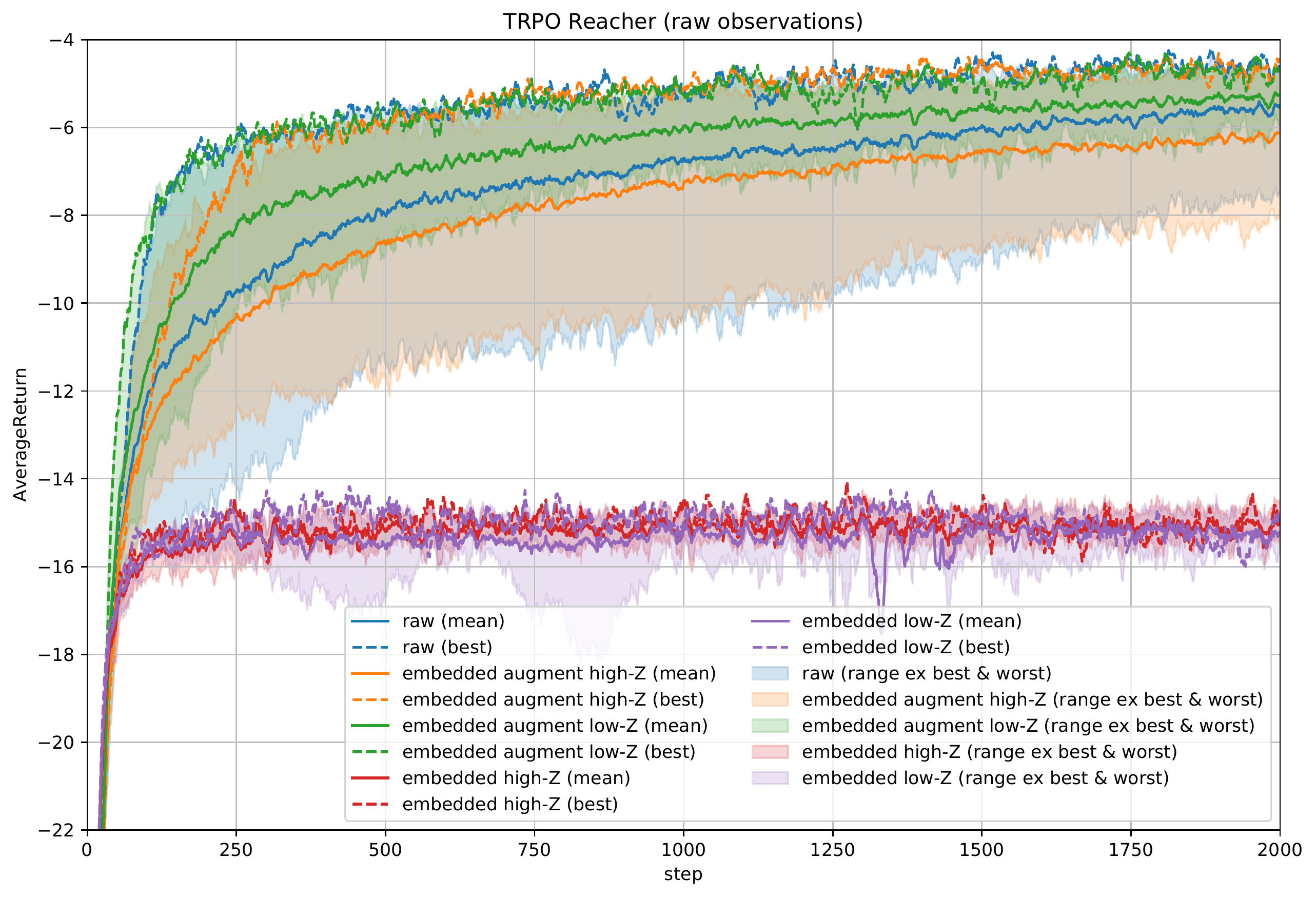}
\caption{
Training curves for the cartpole and reacher environments using trust region policy optimization (TRPO).
Solid lines show the mean value of smoothed performance across 15 random seeds.  Shaded regions show the range from the second worst performing curve to the second best at each training step.  Dashed lines show the smoothed performance of the training run that reached the best performance at any point in training.  High-Z curves used a 10 dimensional embedding space, while low-Z used a 3 dimensional embedding space. (Both cartpole and reacher have 4 dimensional raw state spaces.)  For 'embedded augment' curves the RL agent observes both the robot state and the embedding, 'embedded' curves observed only the embedding space.
While in general the embedding spaces achieve similar best case performance with raw states (in blue), we see significant improvements in reduced seed variance, particularly with the low-Z embeddings (green).
}
\label{fig:trpo_train}
\end{center}
\vspace{-15pt}
\end{figure*}

\begin{figure*}
\begin{center}
\includegraphics[width=0.5\linewidth]{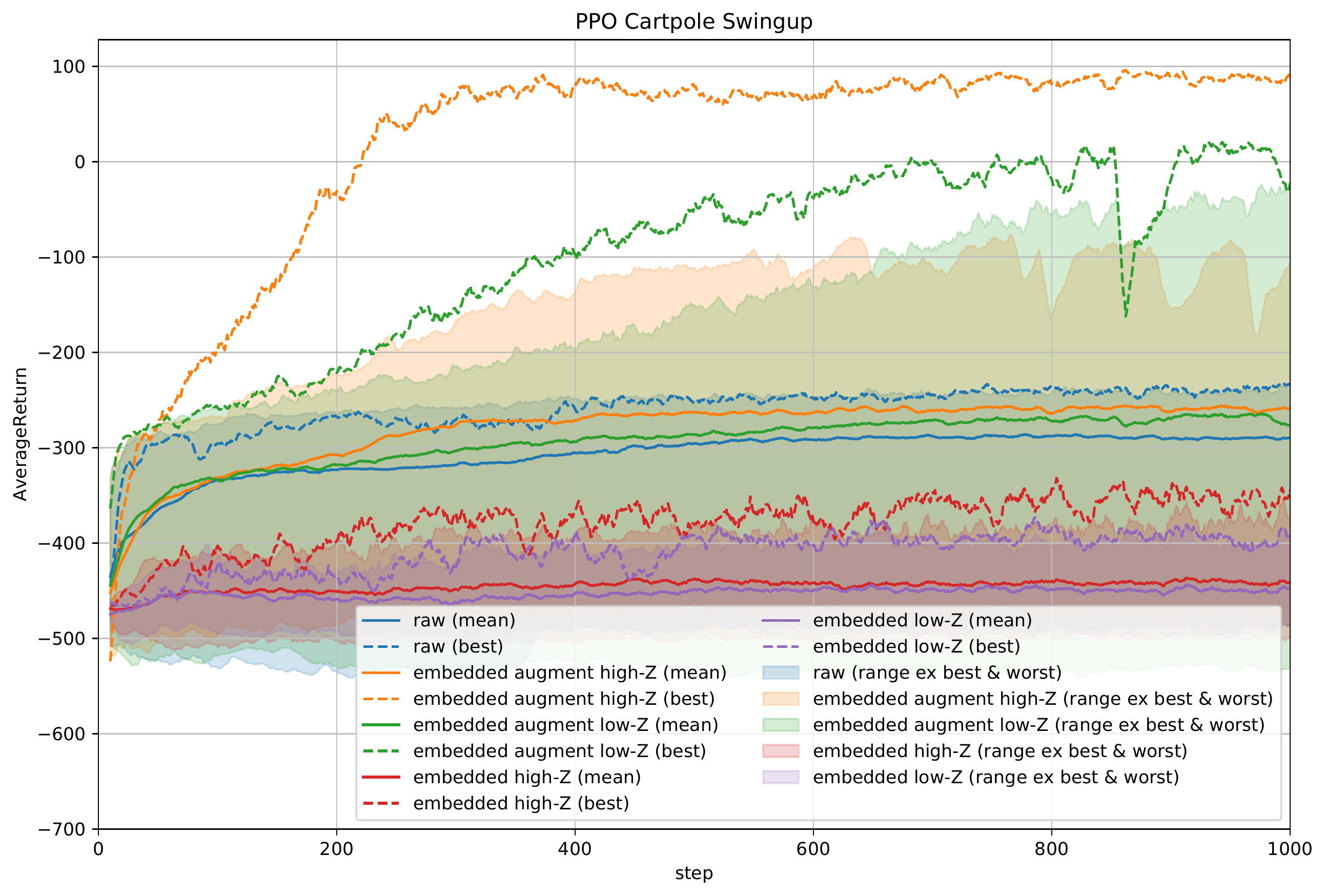}%
\includegraphics[width=0.5\linewidth]{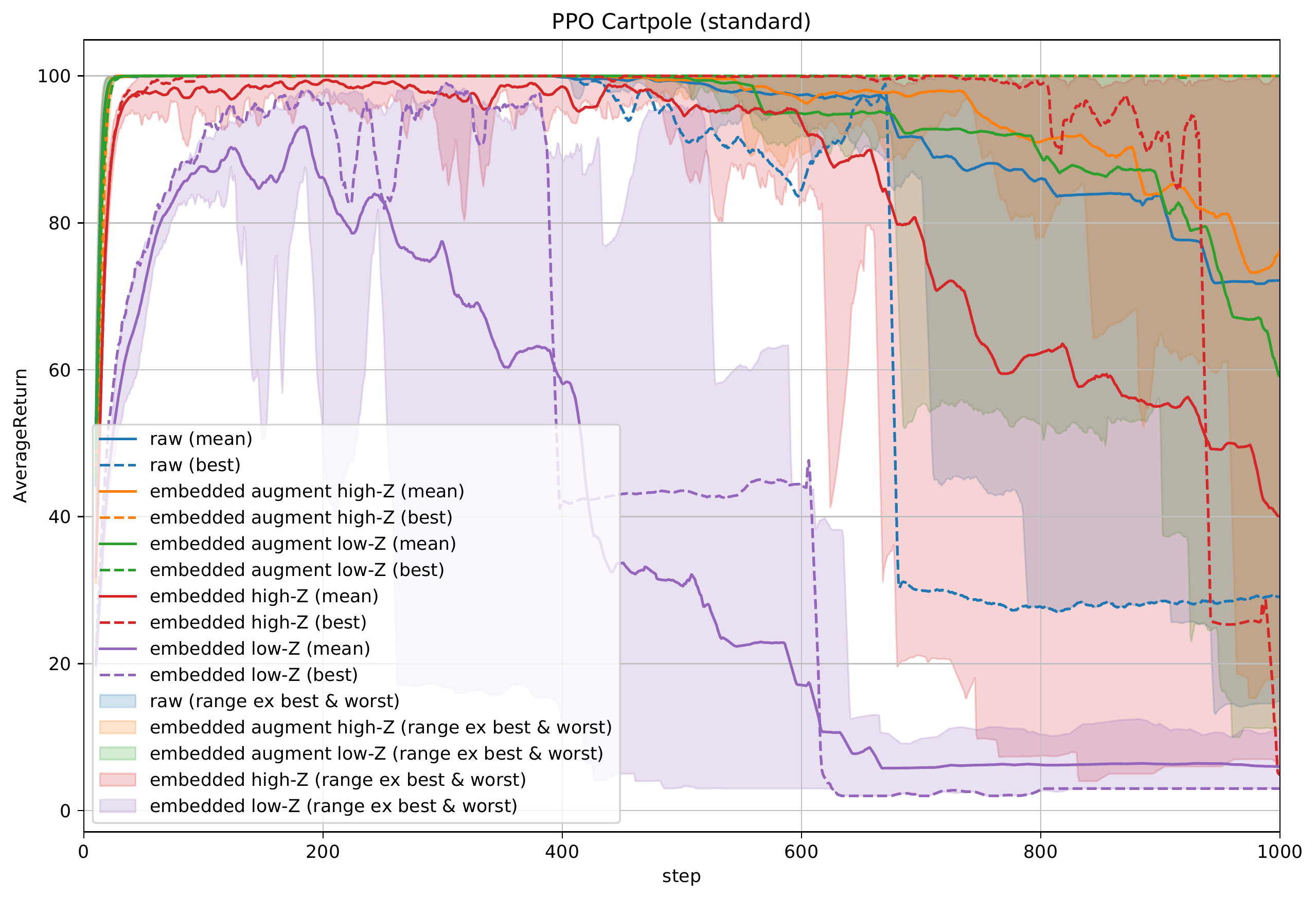}
\includegraphics[width=0.5\linewidth]{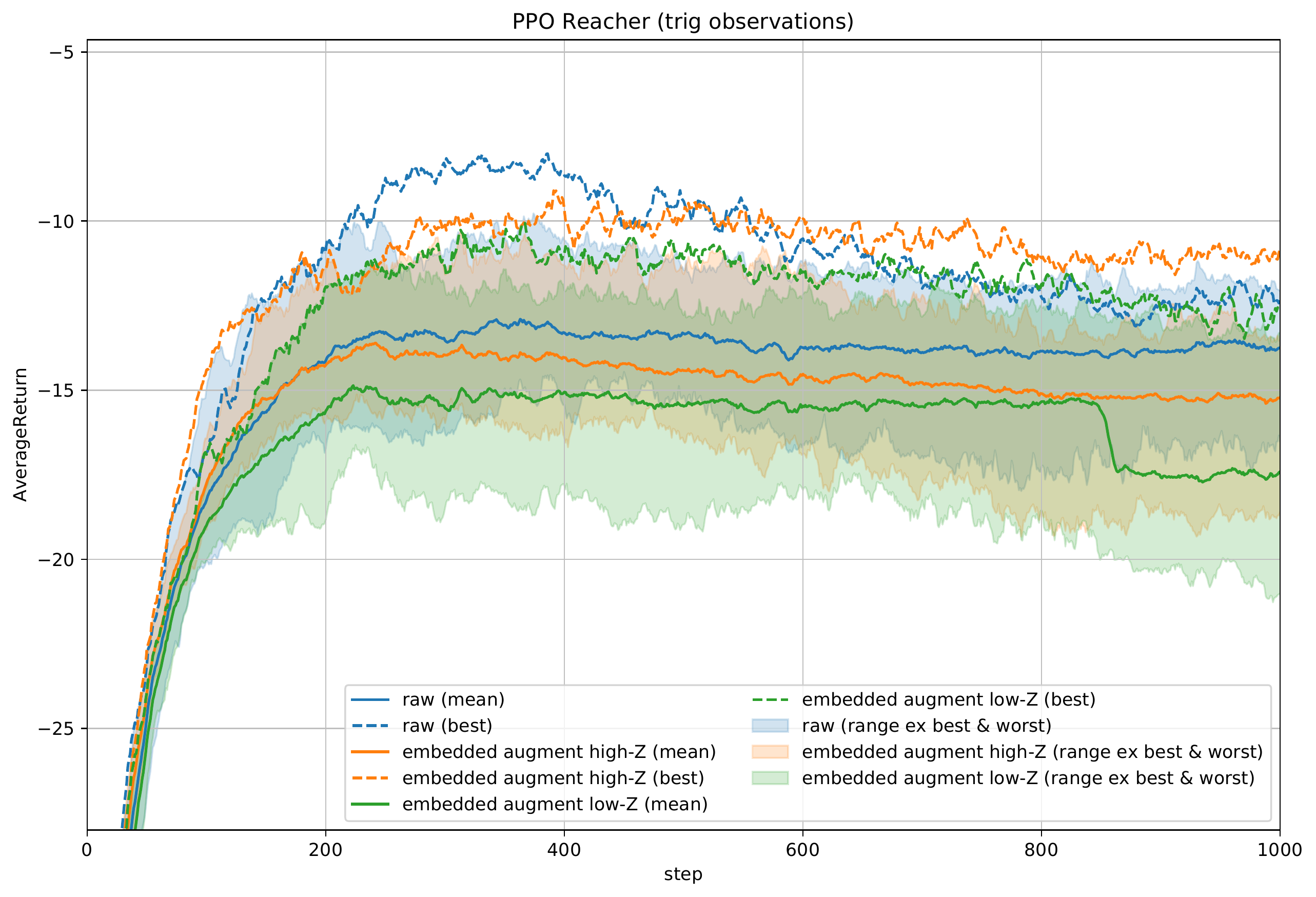}%
\includegraphics[width=0.5\linewidth]{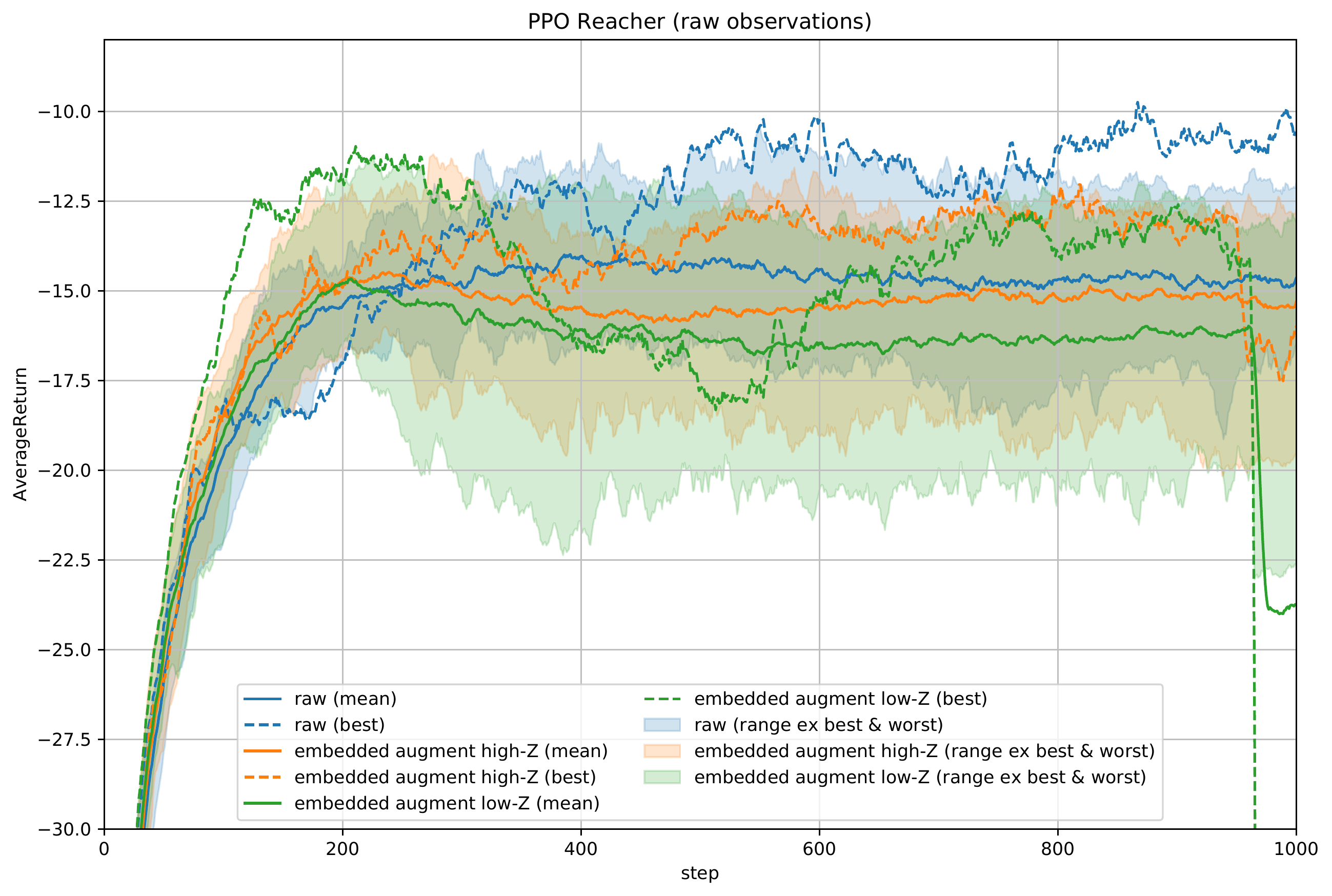}
\caption{
Training curves for our cartpole and reacher environments using proximal policy optimization (PPO).  
Conventions follow Figure \ref{fig:trpo_train}, please see that caption for further detail.
In our experiments PPO generally under-performed TRPO, especially on the reacher where none of the spaces produced particularly strong policies. In the standard cartpole we see similar results to TRPO where most spaces find nearly perfect policies, but the cartpole swingup task shows a situation where our embedding spaces produce a slight improvement in average cases, but substantial improvement in the best cases over learning with raw state values.
}
\label{fig:ppo_train}
\end{center}
\vspace{-10pt}
\end{figure*}

We examine the performance benefit of using our algorithm with two reinforcement learning algorithms, trust region policy optimization (TRPO) and proximal policy optimization (PPO).  We use the reference implementation of these algorithms provided by the garage project \cite{garage}. TRPO and PPO are both policy gradient algorithms, but with different underlying mechanisms.  Policy representations were parameterized as a neural network with two fully connected hidden layers of 32 units.  Experiments were run with a discount factor of 0.99 for 100 time steps, except for the cartpole swingup, which got 200 time steps (to allow enough time to swing up and demonstrate balance of the pole).

Figure \ref{fig:trpo_train} shows training curves for TRPO and Figure \ref{fig:ppo_train} shows the same curves for PPO.  All of the plots show average (un-discounted) performance across the training runs.  The cartpole swingup task in particular has some observable qualitative performance levels: policies that achieve near 100 can quickly swing up the pole from any initial state and maintain stable balance for the rest of the episode.  In the 0 to -100 range are policies that can swing up the pole but fail to achieve stable balance.  The -200 to -300 range are policies that are putting energy in the pole, but not in a controlled way.  In the -600 range are policies that have learned to zero controls to avoid the control penalty and do nothing.  For the reacher task it can generally be said that scores around -6 are effectively achieving the task of reaching to the target point, where as scores below -14 do not appear to demonstrate significant goal directed behavior.  

In the standard cartpole task (where the pole starts in a nearly upright position with low velocity) the baselines are extremely strong, achieving near perfect performance almost immediately.  In these cases augmenting with the embedded state seems to be even or slightly destabilizing. Using the embedded state only can still achieve good performance, especially with TRPO, though not as good as baseline.

In the cartpole swingup task, augmenting with the embedded state offers significant performance improvement in both TRPO and PPO.  In TRPO we see slightly faster rise time, better average performance from both high-Z and low-Z embeddings, similar best case performance, and reduced seed variance towards the end.  With PPO we see substantially better best and 2nd best case performance from both embeddings, and slightly improved average performance, though PPO is generally under-performing TRPO in our experiments.

For our reacher tasks with TRPO, we see much smaller variations.  With trigonometric observations the baseline reacher appears difficult to outperform, and our low-Z embedding is even, with the high-Z underperforming early and then catching up.  With angular (raw) observations the low-Z embedding offers some early benefits in reduced variance, though high-Z slightly under performs.  With PPO both embeddings appear to slightly under perform, and best case results are mixed. Absolute performance on the PPO reacher tasks is not good compared to TRPO though.

When training with policies that observe only the embedding we generally do not get good performance, which suggests that these state embedding are not encoding enough information to fully observe the system state, or that that learning problem is particularly difficult.  In the case of the cartpole clearly some system state can be inferred though due to respectable (though sub-baseline) performance on the standard cartpole task.  Also, as noted earlier, the reacher spaces have a changing goal position that is not observed by our embedding only experiments, so we do not expect to see good performance on this task. 

Overall, we see a couple of patterns in performance when augmenting with embedding spaces.  On tasks where the baseline algorithm is already very strong and reliable we don't see improvements, and we think this is reasonable as there may not be much room for improvement.  The standard cartpole is firmly in this category, and trigonometric observations of the reacher represent a very engineered feature space, which may be difficult to improve upon.  

In tasks where the baseline can achieve good performance but does not do so reliably is where we see the most improvement, which clearly describes the cartpole swingup, and to a lesser extent the reacher with angular (raw) observations.

\section{Conclusion \& Discussion}

In this paper we've proposed a new form of state embedding to learn the geometry of the plan-space of an environment from expert demonstrations, in this case supplied by a motion planner.  We've shown how to train these embeddings, and that once trained they appear to improve the performance and reliability of policy gradient RL algorithms when used to augment the observed state space.

In future work we hope to examine the impact on a greater variety of RL algorithms as well as looking at more varied problem spaces.
The experiments in this paper have formulated our embedding from an already observed joint state space, however it could also be formulated to be a function of a more difficult to interpret observation space such as images, where it might also offer particular advantages in encoding information about the environment beyond robot joint states.

Another interesting direction of future work could be to evaluate the usefulness of this technique using human demonstrations, this could be particularly relevant to complex problems where effective planners are not readily available.

The performance improvements we have observed in RL problems suggest the need for a better understanding of the reward contours of learned policy functions.  Since our embedding space is a function of the observed state space in these example problems, it could be interpreted as an expansion of the expressiveness of the policy network through a pretrained sub-network.  Although we offer some intuition as to why our embedding objective should produce a useful embedding function, the field lacks a more principled understanding of the need for expressiveness in policy functions.

\bibliographystyle{IEEEtran}
\bibliography{references}

\end{document}